%% file: main.tex
\documentclass[runningheads]{llncs}

\usepackage{eccv}

\usepackage{eccvabbrv}

\usepackage{graphicx}
\usepackage{booktabs}

\usepackage[accsupp]{axessibility}  %

\usepackage{hyperref}

\usepackage{orcidlink}

\usepackage{amsfonts}
\usepackage{amssymb}
\usepackage{amsmath}
\usepackage{calc}
\usepackage{caption}
\usepackage{multirow}
\usepackage{multicol}
\usepackage{pifont}
\usepackage[table,x11names,dvipsnames]{xcolor}
\usepackage{xcolor}
\usepackage{pgfplots}
\pgfplotsset{compat=1.18}
\usetikzlibrary{pgfplots.groupplots}
\usetikzlibrary{fillbetween}
\usepackage[normalem]{ulem}
\usepackage{array}
\usepackage{makecell}
\usepackage{wrapfig}

\newcommand{\cmark}{\textcolor{ForestGreen}{\ding{51}}}
\newcommand{\xmark}{\textcolor{BrickRed}{\ding{55}}}
\newcommand{\pgheader}[1]{\noindent \textbf{#1}}
\newcommand{\mT}{\mathcal{T}}
\definecolor{LightGrey}{rgb}{0.8,0.8,0.8}

\begin{document}
\title{Whareformer: Learning to Track What is Where in Long Egocentric Videos}

\author{Jacob Chalk\inst{1} \quad\quad
Saptarshi Sinha\inst{1} \quad\quad
Dima Damen\inst{1} \\Yannis Kalantidis\inst{2} \quad\quad Diane Larlus\inst{2}}

\authorrunning{J.~Chalk et al.}

\institute{University of Bristol, UK \and
NAVER LABS Europe, France \\ \url{https://jacobchalk.github.io/Whareformer/}}

\maketitle

\input{sec/0_abstract}    
\input{sec/1_introduction}
\input{sec/2_related_work}

\input{sec/3_whareformer}
\input{sec/4_experiments}
\input{sec/5_conclusion}

\bibliographystyle{splncs04}
\bibliography{main}

\input{sec/X_suppl}

\end{document}

%% file: sec/0_abstract.tex
\begin{abstract}
The recently established `Out of Sight, Not out of Mind' (OSNOM) task for egocentric videos focuses on tracking objects that are moved by the camera wearer, online, maintaining knowledge of instance locations throughout the video even when they leave the field of view or become heavily occluded.
In this paper, we propose the first learning-based solution to the OSNOM task: Whareformer, a transformer-based model with two components: an updatable memory of established tracks and a track assignment module that associates observations with existing tracks in a feed-forward manner.
Whareformer jointly reasons over evolving object appearance (\textit{what}) and updated 3D location (\textit{where}), and employs a dedicated New Track token to reason about novel objects.

Thanks to its design choices of using relative distances and evolving track representations, Whareformer is trained on a small set of 56 videos but achieves SOTA performance on 260 long test videos from three datasets: EPIC-KITCHENS-100 (unseen videos), IT3DEgo, and HD-EPIC,  with significant absolute improvements over prior work.
\end{abstract}

%% file: sec/1_introduction.tex
\begin{figure}[t]
\centering
\includegraphics[width=\linewidth]{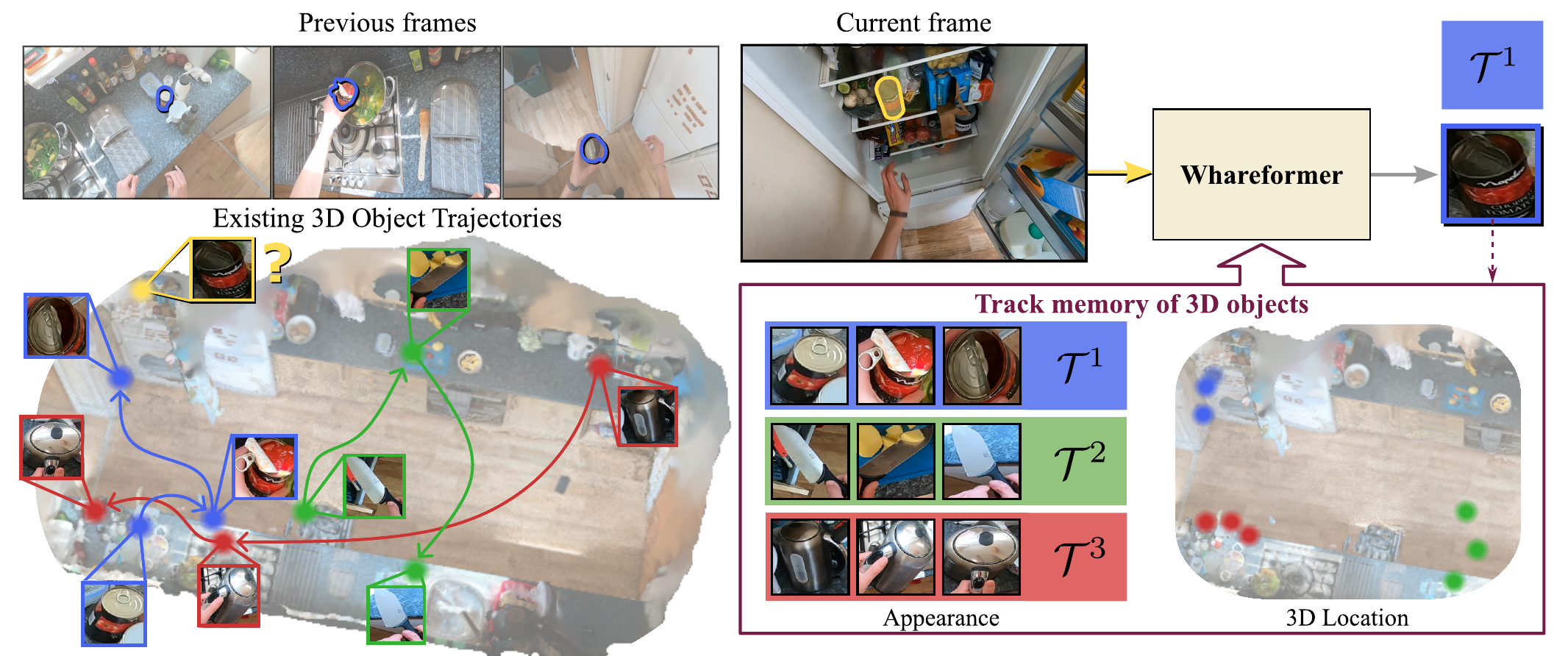}
\caption{\textbf{Overview of Whareformer.} In the OSNOM task, the goal is to assign the current observation (highlighted in \textcolor{Dandelion}{\bf yellow} in the current frame) of a 3D egocentric scene with one of the known objects (which respectively correspond to a \textit{kettle} in \textcolor{red}{\bf red}, a \textit{knife} in \textcolor{ForestGreen}{\bf green}, or a \textit{tin of chopped tomatoes} in \textcolor{NavyBlue}{\bf blue}). We propose Whareformer, a model that jointly reasons about the appearance and the location of objects to decide whether to associate the observation with an existing track or initialise a new one.} 
\label{fig:teaser-whareformer}
\end{figure}

\section{Introduction}
\label{sec:intro}

Humans effortlessly maintain a mental map of their surroundings, simultaneously tracking multiple objects of interest, such as the keys on the counter, the phone on the coffee table, the mug by the sink. This ability to maintain an understanding of ``what is where'' for multiple objects, even when they are temporarily out of sight, is a core component of how humans interact with the world. For an autonomous agent to perform complex tasks in a human environment, replicating this capacity is a step towards human-like spatial intelligence. While this spatial understanding could be learned implicitly, representing it as an explicit 3D map provides a stable, viewpoint-invariant basis for locating and interacting with objects within this environment.

To formalise this concept for egocentric videos, Plizzari \etal introduced the \emph{Out of Sight, Not Out of Mind} (OSNOM) task~\cite{Plizzari2025OSNOM}. 
It measures a model's ability to (1) recall the 3D locations of objects over time, even when out of view due to egomotion, object movement, or occlusion and (2) maintain the identity of objects as they change their appearance due to interactions and manipulations.
The task can be simplified by explicit 3D reasoning, particularly when an object is occluded or when egomotion observes the same object from different viewpoints.
An effective method would (1) maintain a coherent understanding of \emph{what is where} within a consistent 3D coordinate frame, mirroring the spatial cognitive ability of humans and (2) learn how object appearances change during interactions, enabling accurate tracking for objects that are closely positioned.

OSNOM focuses on unstructured, human-centric egocentric scenarios characterised by rapid camera motion, frequent occlusions, and the need to maintain object permanence over extended timescales (typically up to half an hour). Prior work~\cite{zhao2024instance, hao2024ego3dt, Plizzari2025OSNOM, bhalgat20243d} tackles these challenges using heuristic association strategies based on fixed, manually defined thresholds. This restricts adaptability, forcing associations even when ambiguous, and limits the ability to reason about when to initialise or retain tracks. 

Departing from these approaches,
we introduce a method that combines two trainable modules, one for track candidate selection and one for track assignment. \Cref{fig:teaser-whareformer} illustrates how our method jointly leverages 2D appearance similarities (\emph{what}) and the metric distance between objects lifted to 3D (\emph{where}) through a learned What-and-Where Transformer (aka \emph{Whareformer}). We efficiently store track representations in memory via adaptive clustering of evolving object-level appearances and changing locations. Importantly, new track creation is treated as a first-class decision through an explicit token, rather than being triggered only when a manually defined cost threshold is exceeded. Through learning, the model dynamically handles complex scenarios (such as occlusion, objects moving to locations where others were removed, or objects reappearing from a different viewpoint) and learns when to initialise new tracks for novel observations, with additional gains possible through training with DAgger correction~\cite{ross2011dagger}. Taken together, these designs yield a robust and efficient egocentric object tracker.

\pgheader{Contributions.}  
We propose \textit{Whareformer}, a novel architecture for long-term 3D object tracking in egocentric videos.
We train \textit{Whareformer} on labelled trajectories of active objects on a split from the EPIC-KITCHENS-100 dataset.
Learnt directly from data, \textit{Whareformer} considers the relative distances of observations to established tracks, and predicts track assignments, including the decision to start a new track via an explicit token.
As \textit{Whareformer} captures \textit{relative} appearance and location, it can generalise to unseen objects and environments, including unseen datasets. This is demonstrated by performance improvements over three egocentric benchmarks~\cite{Damen2020RESCALING,zhao2024instance,perrett2025hdepic}.

%% file: sec/2_related_work.tex
\section{Related Work}
\label{sec:related_work}

\noindent \textbf{2D tracking in egocentric videos} faces challenges due to the videos' rapid viewpoint shifts and the objects' large apparent motion and frequent occlusions.
This requires specialised designs over general trackers~\cite{huang2023tracking, dunnhofer2023visual, manigrasso2024online, goletto2024amego, dunnhofer2025tracking}, that focus on morphing objects~\cite{huang2023tracking}, interaction memories~\cite{manigrasso2024online, goletto2024amego, tang2023egotracks} or cross-view association~\cite{ardeshir2018integrating, han2021multiple}. Despite these advances, such methods suffer from the inherent limitations of 2D trackers, which struggle to recover objects that have not been observed for minutes, as their viewpoint and hence their appearance could have drastically changed.
In contrast to these methods, Whareformer enables tracking across occlusions, extended periods of absence, and viewpoint changes thanks to its embedded 3D reasoning.

\pgheader{3D tracking in egocentric videos} leverages 3D information in order to ground objects in the physical world. Early efforts demonstrated that visual odometry can help overcome 2D tracking shortcomings for people in egocentric cameras~\cite{alletto2015egocentric}. Recent benchmarks, such as HOT3D~\cite{banerjee2025hot3d}, have further enabled 6DoF tracking of rigid objects using multi-view headset data. While effective during active manipulation, these methods assume near continuous object visibility or specialised hardware. To achieve object permanence in a larger scene and for long videos, recent methods~\cite{zhao2024instance, hao2024ego3dt, Plizzari2025OSNOM, bhalgat20243d} build upon static 3D object localisation techniques~\cite{achlioptas2020referit3d, chen2020scanrefer, avetisyan2024scenescript, arnaud2025locate, mai2023egoloc, qian2023understanding}. These frameworks address the challenges in 3D tracking through diverse strategies: IT3DEgo~\cite{zhao2024instance} lifts 2D objects into 3D to improve temporal consistency, while Ego3DT~\cite{hao2024ego3dt} reconstructs geometry using a pre-trained 3D model and performs hierarchical association.
The Lift, Match, and Keep (LMK) approach in~\cite{Plizzari2025OSNOM}, explicitly formalises the spatial cognition necessary to track active objects even when they are out of sight. 
This method reconstructs the static scene and associates observations using a hand-crafted appearance-location cost and the Hungarian algorithm~\cite{munkres1957algorithms}. 

In contrast with all above works, Whareformer proposes trainable modules that jointly reason over appearance and 3D cues. These modules learn to reidentify objects even from different viewpoints, and dynamically infer the need for a new track via an explicit token. This yields improved performance and generalises across egocentric datasets.

Whareformer is related to works in general 3D tracking~\cite{wang2022deepfusionmot, weng2020ab3dmot, yin2021center, stearns2022spot,rajasegaran2022tracking} where transformer-based models have been used for track association. These works focus on short driving sequences (a few minutes) where objects maintain their appearance and typically utilise explicit 3D sensors like LIDAR.
Similar to these works, Whareformer learns how to track directly from data.
Differently from the scenario they tackle, egocentric videos require
unique reasoning about changing appearance and location of general objects including objects that come together at very close proximity (e.g. a spoon goes inside the jar) as well as dealing with extended periods of non-observations (for many minutes).

%% file: sec/3_whareformer.tex
\section{Whareformer}
\label{sec:whareformer}

\begin{figure}[t!]
    \centering
    \includegraphics[width=\linewidth]{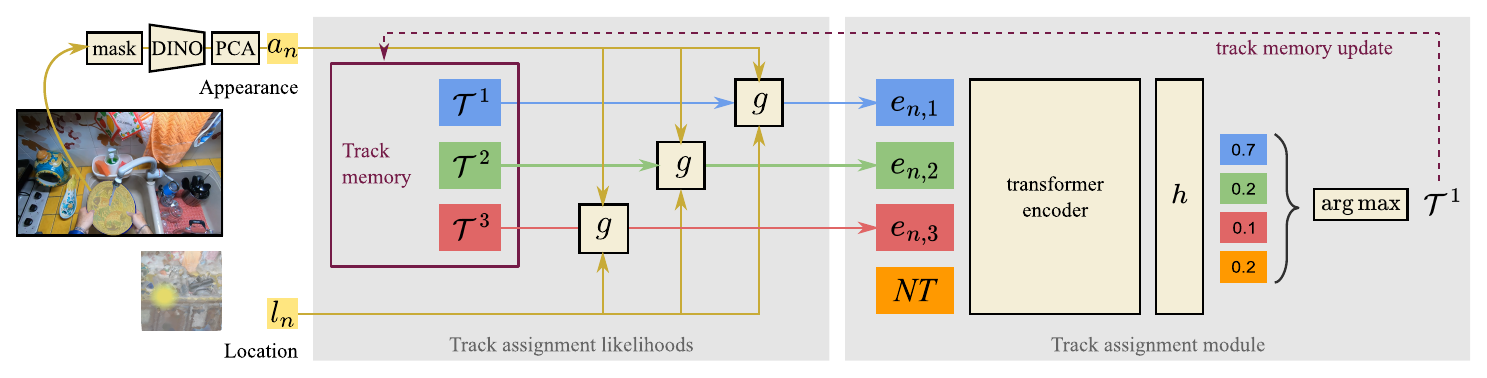}
    \caption{\textbf{Whareformer architecture.} The current observation is represented by appearance $a_n$ and location $l_n$ descriptors (\Cref{subsec:representing}), which are fed into our first module (\Cref{subsec:assignment}). Using a memory of tracks constructed so far (\Cref{subsec:tracks}), it produces embeddings representing the likelihood that this observation corresponds to each track. These embeddings, combined with a new track ($\mathrm{NT}$) token, are fed to a transformer (\Cref{subsec:assignment}) that produces the final track assignment. This prediction is used to update the track memory (\Cref{subsec:track_updates}).}
    \label{fig:architecture}
\end{figure}

In this section, we present our method 
for the OSNOM task. First, we formally define the problem (\Cref{subsec:prelim}). Then, we describe the representations used to encode the appearance and 3D location of observed objects (\Cref{subsec:representing}) and of tracks (\Cref{subsec:tracks}). Next, we describe the different components of \textbf{Whareformer}, as illustrated in \Cref{fig:architecture}: the computation of \textit{track assignment likelihoods} that compare an observation to all tracks constructed so far (\Cref{subsec:selection}), the \textit{track assignment module} (\Cref{subsec:assignment}) that chooses which track to assign that observation to  and the \textit{track memory update} mechanism that alters the track memory using the current observation and its track assignment  (\Cref{subsec:track_updates}). Finally, we briefly discuss training and inference (\Cref{subsec:training}).

\subsection{Problem formulation} 
\label{subsec:prelim}
Given an egocentric video with $F$ total frames, let $\mathcal{O} = \{ o_{1}, o_{2}, \dots, o_{N} \}$ denote a temporally ordered set of object observations corresponding to $M$ unique objects, where multiple observations may occur within the same frame, objects are only visible in a subset of frames, and may be observed multiple times throughout the sequence. Each observation is defined as $o^{t}_{n} = (a^{t}_{n}, l^{t}_{n})$, consisting of an appearance descriptor $a^{t}_{n}$ and a 3D location $l^{t}_{n}$, both derived from a corresponding segmentation mask at frame $t$.

The goal is to recover consistent 3D trajectories for all objects by assigning the $N$ observations to $K$ tracks $\mathcal{T} = \{ \mathcal{T}_{1}, \mathcal{T}_{2}, \dots, \mathcal{T}_{K} \}$, where $K \ll N$. Each track $\mathcal{T}_{k}$ is represented by an appearance representation $A(\mathcal{T}_{k})$ and a location representation $L(\mathcal{T}_{k})$, which are compared with $a_{n}$ and $l_{n}$ to associate each object observation $o_{n}$ with the corresponding track for that unique object instance.

For each object $m$, the corresponding track $\mathcal{T}_{k}$ must maintain an accurate estimate of its 3D position $l^{t}_{m}$ across all frames $t \in \{1, \dots, F\}$, including when the object is not visible, thereby assessing the method's ability to reason about object permanence over long temporal horizons.

\subsection{Representing observations}
\label{subsec:representing}

For the visual appearance, we extract high-dimensional features using a pre-trained DINOv2 model~\cite{oquab2024dinov}. To isolate the object's visual content from its surrounding context, we use the object's mask~\cite{khosla2025relocate}. 

To enhance discriminative power and reduce redundancy, we perform dimensionality reduction using a Principal Component Analysis (PCA) projection trained on all visual features from the training set (see~\Cref{subsec:datasets-whareformer}). The final feature $a_{n}$ is obtained by first applying $\ell_{2}$ normalisation to the DINOv2 feature, projecting it through the learned PCA transformation, and then applying $\ell_{2}$ normalisation once more to the reduced feature. This produces the compact $D$-dimensional descriptor we refer to as $a_{n}$. 

For the 3D location, we obtain location estimates following the pipeline described in~\cite{Plizzari2025OSNOM}, which aligns monocular depth with 3D scene reconstructions and lifts the centre of the objects into the world coordinate frame using this aligned depth, leading to the location descriptor $l_n$.
Using the same lifting approach as~\cite{Plizzari2025OSNOM} allows fair comparison, but we ablate removing the alignment step and show our approach is robust to how the 3D location is acquired.

\subsection{Representing object tracks}
\label{subsec:tracks}

Accurate long-term association requires a robust and evolving representation for each track $\mT_{k}$ that effectively summarises the corresponding object's history of visual appearance (\textit{what}) and 3D location (\textit{where}). 

\pgheader{Track appearance.} The object's appearance evolves over time due to manipulation, interactions, occlusion as well as viewpoint and illumination changes. An online representation must capture this evolving history efficiently. However, a careful design is required: a short sliding temporal window cannot recall distant appearances whereas an infinite buffer is intractable and 
sensitive to outliers. The chosen representation should therefore balance context with efficiency.

We address this challenge by treating appearance as an infinite data stream, clustered online via the DenStream algorithm~\cite{cao2006density}. It efficiently approximates an infinite buffer of appearance descriptors using two sets of temporally weighted clusters: persistent clusters that ensure long-term consistency, and transient clusters that adapt to short-term changes.
This leads to the definition of the track's appearance $A(\mT_{k}, t)$ as the combination of both types of clusters at time $t$. It is computationally tractable and memory-efficient (see the supplementary material for details on the DenStream algorithm). 

\pgheader{Track location.}
The location history of a track must be robust and discriminative of where the object has been seen last. Using a track's entire location history is undesirable, as paths overlap across multiple objects, creating ambiguity as objects move in the environment.

A short history is still desirable to handle noisy estimates but also objects moving around space during interactions.
We thus represent $L(\mT_{k}, t)$ as a temporal buffer containing the last $W$ 3D locations assigned to the track, with $W$ chosen to cover about one second of video.

\subsection{Learnable track assignment likelihoods} 
\label{subsec:selection}

For each observation $o_{n}$ in the current frame at time $t$,
we need to decide whether to assign it to an existing track or create a new one.
Let $\mT_{k}$ denote one of the existing tracks, 
we define the initial likelihood of assigning $o_{n}$ to track $\mT_{k}$ as the combination of appearance and location distance measures respectively:

\begin{equation}
    c_{n,k}^{A} = \min_{a' \in A(\mT_{k}, t)} \text{dist}(a_{n}, a')^{2},\quad 
    c_{n,k}^{L} = \min_{l' \in L(\mT_{k}, t)} \text{dist}(l_{n}, l'),
\label{eq:distances}
\end{equation}
where $\text{dist}(\cdot,\cdot)$ denotes the Euclidean distance. These two terms are concatenated and embedded into a high-dimensional space using a linear projection $g$:

\begin{equation}
e_{n,k} = g([c_{n,k}^A, c_{n,k}^L]).
\label{eq:trackpotentials}
\end{equation}

The embedding vector $e_{n,k} \in \mathbb{R}^d$ could be understood as closely related to the likelihood that observation $o_{n}$ corresponds to track $\mT_{k}$. 
It jointly encodes appearance and spatial information and is trained to model their interaction, thanks to the learnable projection $g$. As shown in~\Cref{sec:experiments}, this design not only improves performance but also 
fosters generalisation to other objects or environments, as it encodes distances rather than actual features.

Next, we describe how these embedding vectors are used as input tokens to a transformer that performs the final matching between observations and tracks.

\subsection{Learnable assignment module} 
\label{subsec:assignment}

In this section, we describe the transformer-based module used to assign observations to tracks in a principled and learnable manner. This module jointly weighs appearance and location cues, directly producing a probability distribution over all existing tracks as well as a potential \textit{new track}.

\pgheader{New track token.}
To account for observations that do not match any existing tracks, we prepend a New Track token $\mathrm{NT} \in \mathbb{R}^{d}$ to each sequence of embedded tokens, inspired by the `$\mathrm{[NT]}$' token in the architecture proposed in~\cite{price2022unweavenet}. 
This $\mathrm{NT}$ token enhances the model with the ability to decide whether or not to create a new track and add it to the track memory.

This is distinct from heuristic methods commonly used in prior work~\cite{Plizzari2025OSNOM,zhao2024instance}, which treat new track creation as an implicit failure case (\eg when all match costs exceed a threshold~\cite{Plizzari2025OSNOM}). Instead, the $\mathrm{NT}$ token makes it an explicit choice that the model can learn to select.
This decision can be directly weighed against all other potential assignments.

\pgheader{Input sequence.} 
We prepend the $\mathrm{NT}$ token to the embedded likelihood vectors constructed in the previous section.
The input sequence for observation $o_n$ is

\begin{equation}
  \mathbf{S}_{n} = [\mathrm{NT},\:e_{n,1},\: e_{n,2},\:\ldots,\:e_{n,T}] \in \mathbb{R}^{(T+1) \times d},
  \label{eq:transformer_input_sequence}
\end{equation}
where $T$ is the number of tracks created so far, \ie at time $t$. 

As the set of existing tracks has no intrinsic order, we treat the embedded tokens $\mathbf{e}_{n,k}$ as an unordered set and do not add positional encodings, ensuring the model is permutation-invariant. Each sequence $\mathbf{S}_{n}$ is passed through a transformer encoder, allowing all tokens (each token representing a potential assignment) to attend to one another. This contextual attention lets the model evaluate the suitability of assigning an observation to a track relative to all other tracks and to the option of creating a new one, enabling robust decisions. 
The output is a contextually-refined representation $\mathbf{Z}_n$.

\pgheader{Track assignment.} To make the final assignment decision, a linear classification head $h(\cdot)$ is applied to each of the $T+1$ output tokens, mapping them to a set of logits that represent the final assignment scores:

\begin{equation}
  \hat{y}_{n} = \text{softmax}\left(h(\mathbf{Z}_{n})\right) \in \mathbb{R}^{T+1}
\end{equation}

$\hat{y}_{n}$ represents a probability distribution over assignment to any existing track or to a new one.
The final decision is the argmax of this distribution, \emph{unifying the track matching and creation decisions}
within a single forward pass.

\subsection{Updating the memory of tracks}
\label{subsec:track_updates}

Once the current observation is assigned to an existing track or a new one, we can update that track's appearance and location representations. For existing tracks, the observation's location is appended to the history buffer, discarding the oldest entry if it exceeds the maximum buffer size, whilst its appearance is given as input to the DenStream clustering algorithm, triggering an update of the clusters. For new tracks, the representation is initialised with the observation's appearance (creating a first cluster) and location (starting a buffer of size~1).

\subsection{Training and inference}
\label{subsec:training}

Whareformer is trained using teacher forcing with ground truth object track IDs. A training sample considers an observation at time $t$ along with a memory formed from ground-truth tracks up to time $t-1$. The correct assignment, to an existing or new track, for this observation is used as the training signal. Since inference relies on predicted track assignments rather than ground-truth tracks, this introduces an inference distribution shift. We ablate the impact of this shift by training using DAgger-style correction~\cite{ross2011dagger} in~\Cref{subsec:ablations}.

\pgheader{Training details.} 
Batches are created by sampling individual observations from any frame from any training video. As the number of existing tracks varies across frames, we pad the input to the maximum track count within the batch and apply masking so that padded entries are ignored during training. The model is trained with a cross-entropy classification loss between predicted track assignments $\hat{y}$ and ground-truth labels $y$.

\pgheader{Inference.} During online tracking, for each new observation, we compute the track assignment likelihood as defined in~\Cref{eq:trackpotentials}, construct the transformer input sequence following~\Cref{eq:transformer_input_sequence}, and feed it to the transformer model, treating the number of observations as the batch dimension.

Since assignments are predicted independently per observation, conflicts may arise when multiple observations select the same track.  
We resolve conflicts iteratively, accepting the highest-confident assignment, then assigning other observations to alternative tracks or a new track based on confidence.
This approach is faster than the Hungarian matching algorithm as each observation in the conflict is revisited once.

%% file: sec/4_experiments.tex
\section{Experiments}
\label{sec:experiments}

\subsection{Experimental details}
\label{subsec:datasets-whareformer}

Our model is trained
on the EPIC-KITCHENS dataset~\cite{Damen2020RESCALING}, which we divide into a training and test split. 
We assess the cross-dataset generalisation by testing on two other datasets~\cite{zhao2024instance,perrett2025hdepic}. We detail them below.

\pgheader{EPIC-KITCHENS~\cite{Damen2020RESCALING}} is a large-scale egocentric video dataset containing 700 kitchen recordings using a GoPro. We adopt the subset of 110 videos across 45 kitchens defined in the benchmark of~\cite{Plizzari2025OSNOM}, selected for their average length ($\approx$12 minutes) that ensures adequate temporal coverage for object tracking. 

\pgheader{Training/test split.} We use the 110 EPIC-KITCHENS videos selected for the OSNOM task~\cite{Plizzari2025OSNOM}. 
From these, we form a split of 56 training (15 kitchens, 12.2 hours, 13.0 min average length) and 54 test (12 distinct kitchens, 12.2 hours, 13.5 min average length) videos. 

We follow the same evaluation protocol without modifications, using keyframes with at least 3 active objects for evaluation.
The same split is used for ablations and results. 

\pgheader{IT3DEgo~\cite{zhao2024instance}} contains 50 egocentric videos (5.5 min average length) recorded in 10 indoor scenes (kitchen, garage, office, labs) by three participants using HoloLens2. 
This means there is not only a domain shift due to the use of a different recording device (GoPro vs HoloLens) but also a broader range of activities taking place in different environments compared to the training set. 

\pgheader{HD-EPIC~\cite{perrett2025hdepic}} is a validation dataset, recorded using Aria Glasses~\cite{engel2023project}, with 156 kitchen-based egocentric videos (41 hours, 9 kitchens) and 37K object masks (8360 long-term trajectories of unique objects), providing 2D bounding boxes and 3D locations via digital twinning. Evaluating on this dataset is more challenging due to sparse annotations only at pick-up and placement events. 

\pgheader{Implementation details.} We use $D=256$ and $d=64$. DenStream consists of 3 parameters, which we set to $\epsilon=0.25$, $\mu=10.0$, and $\lambda=1\mathrm{e}{-3}$. We train the model for 100 epochs using the AdamW optimiser~\cite{loshchilov2017decoupled} with a learning rate and weight decay of $1\mathrm{e}{-4}$. We leave further details to the supplementary material.

\pgheader{Evaluation.} Our evaluation metrics separately assess the model's performance concerning ``what'' is being tracked and ``where'' it is located. To evaluate the ``where'' (spatial accuracy), we follow~\cite{Plizzari2025OSNOM} and use the Percentage of Correct Locations (PCL) to assess long-term 3D object localisation. 
Inspired by human pose estimation's PCK~\cite{yang2012articulated}, PCL measures 3D localisation accuracy over time, including when objects are out of sight. 
As in~\cite{Plizzari2025OSNOM}, a predicted location is correct if it is within 30cm of the ground truth. 
Results are reported as a single summary metric, the mean PCL (mPCL), representing the average PCL aggregated across all videos and sampled timescales. Additionally, results are visualised using PCL plots, which show how the average PCL across videos varies with time.

To evaluate the ``what'' (identity preservation), we report IDF1~\cite{ristani2016performance} which measures track consistency and determines how effectively the system maintains correct object identities throughout the long and complex egocentric sequences.

\subsection{Compared methods}

We compare to a number of baselines, as well as established tracking methods, which we detail below. All methods use the same masks and 3D locations.

\noindent \textbf{Random Matching} highlights the complexity of the task. Each observation is assigned to either an existing track or creates a new one with uniform probability.

\noindent \textbf{Out of Sight, Lost (OSL)~\cite{Plizzari2025OSNOM}} is a baseline which illustrates how frequently objects enter and leave the field of view: it only tracks objects while they remain visible. Once an object exits the camera's field of view, tracks are terminated and incapable of localising objects in future frames (PCL is zero). If the same object later reappears, a new track is initiated.

\noindent \textbf{Out of Sight, Out of Mind (OSOM)~\cite{Plizzari2025OSNOM}} is an oracle tracker that serves as an upper bound for perfect tracking in the camera coordinate frame. Here, objects are tracked while in-view and frozen otherwise, allowing the track to recover the object when it reappears, but unable to localise it when out-of-view.

\noindent \textbf{ByteTrack~\cite{zhang2022bytetrack}} is a strong 2D multi-object tracking method, operating without 3D knowledge. For our setting, objects are first tracked in 2D and then lifted to 3D to evaluate their 3D localisation capabilities.

\noindent \textbf{IT3DEgo~\cite{zhao2024instance}} is a method for object tracking where single or multiple views of each object are assumed available as a reference for tracking, similar to single object tracking.
We adapt the single-view method such that visual templates are enrolled from the first appearance of each object and subsequently matched across frames using DINOv2~\cite{oquab2024dinov} features. Furthermore, 2D observations are lifted to 3D, while motion is modelled by a Kalman filter assuming piecewise constant velocity, which is reset when large deviations signal object interaction or displacement. Due to the enrolment step in this method, the number of objects is already known, hence it has a strong advantage and comparison is not direct. 

\noindent \textbf{LMK~\cite{Plizzari2025OSNOM}} represents the current state of the art (SOTA) on the OSNOM task, outperforming prior methods~\cite{zhang2022bytetrack, mai2023egoloc, zhao2024instance}. 
LMK computes visual features for appearance and  lifts 2D observations into 3D. These two cues are used to solve data association via the Hungarian algorithm on a cost matrix that blends appearance similarity and 3D distance. Tracks are updated using the matched observations, and new tracks are created only when no valid match is found. 
    
\noindent \textbf{LMK-Inf} is an improved variant of LMK. We equip LMK with \textit{infinite memory} rather than 
only storing the most recent samples.
LMK-Inf retains all past assignments and computes the matching for a new observation against the nearest neighbour in both 3D location and appearance across the track's \textit{entire} history.
While powerful, its ever-growing memory makes it impractical for long videos.

All methods use the same DINOv2 appearance features as Whareformer. Note that we performed a grid search on our training set to tailor LMK's parameters to our training/test split, but found no improvement at inference over the original hyperparameters reported in~\cite{Plizzari2025OSNOM}.

\subsection{Results on the EPIC-KITCHENS dataset}

We evaluate on the EPIC-KITCHENS test split, and report results for all methods in~\Cref{tab:cross_dataset_results} (left).

First, we assess the combined challenge of tracking both ``where'' objects are located and ``what'' their continuous identities are. OSL, which forgets objects immediately after they leave the field of view, exhibits extremely poor localisation (1.2\% mPCL) and track consistency (36.8\% IDF1), reflecting the frequent disappearance and reappearance of objects in egocentric videos.
OSOM cannot localise objects once they leave the camera coordinate frame, thus gets 8.3\% mPCL, while ByteTrack achieves poor localisation (12.3\% mPCL), highlighting the need for 3D reasoning. With 3D information, IT3DEgo shows notable gains in both localisation and identity tracking (40.0\% mPCL and 71.0\% IDF1).
LMK, which combines appearance and location, achieves a clear boost in performance (53.0\% mPCL and 70.0\% IDF1). An additional boost is achieved with our modified version with infinite memory.

Whareformer achieves the best results across all metrics, outperforming all prior methods including SOTA method LMK~\cite{Plizzari2025OSNOM}, by +19.2\% mPCL and +14.0\% IDF1, and also surpassing the infinite-memory variant (LMK-Inf). This represents a significant improvement in both spatial accuracy and identity preservation. 
\Cref{fig:cross_dataset_pcl} (left) also shows that this advantage persists across timescales, from short-term (81.8\% vs.\ 68.5\% at 1 min) to long-term tracking (65.4\% vs.\ 39.9\% at 12 min). These results highlight Whareformer's ability to retain past information as well as reason about the relative location and appearance of objects and tracks to produce improved track assignments.

\input{tables/main_results}

\begin{figure}[t]
    \centering
    \includegraphics[width=\linewidth]{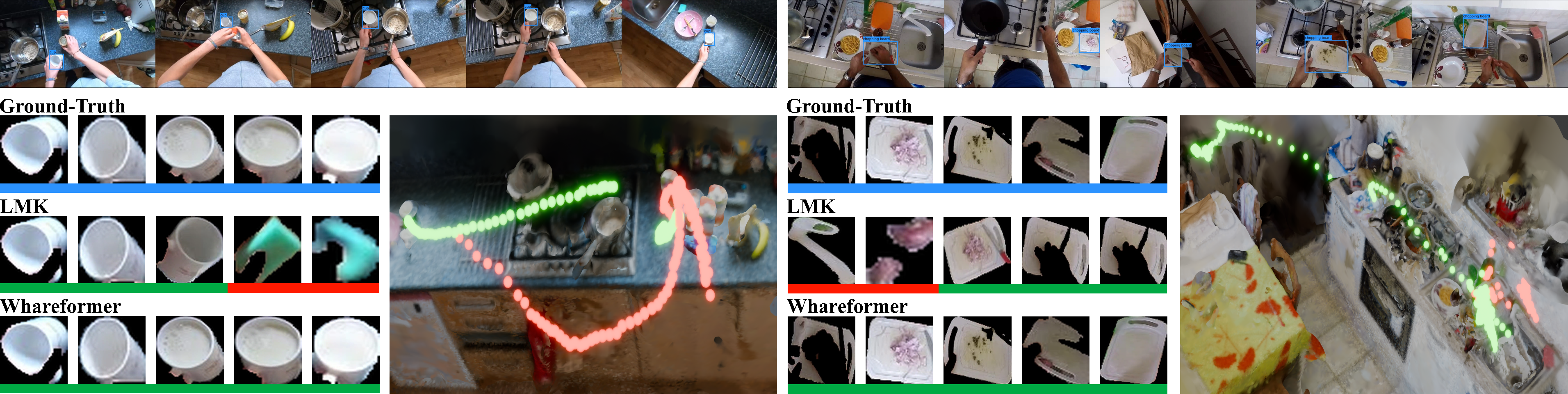}
    \caption{\textbf{Qualitative results.} For each example (left: \textit{mug}, right: \textit{chopping board}), we show the ground-truth (GT) track and the track containing the object's first observation for LMK and Whareformer. For sampled frames along the trajectory we show the corresponding segmented object and 3D trajectory. \textcolor{Green}{\textbf{Green}} underlined crops and 3D orbs denote correct associations with the GT while \textcolor{red}{\textbf{red}} indicate incorrect ones. 
    }\label{fig:wharerformer_qualitative_results}
\end{figure}

\pgheader{Qualitative results.}
We present qualitative results in~\Cref{fig:wharerformer_qualitative_results} for two long-term tracking scenarios.
The first example tracks a mug moved across a hob and then left stationary. LMK initially tracks the mug correctly but loses it to a nearby sponge once it is set down. In contrast, Whareformer maintains the mug's full trajectory with consistent identity.
In the second example, a chopping board is tracked through multiple interactions: from the draining board to a bin to discard trash on it, across the kitchen, then over the hob, and back to the draining board. LMK incorrectly assigns the chopping board to an existing track (initially the tap and later the onion) rather than creating a new track, then briefly tracks the board before losing it during its transit.
The qualitative results show more consistent tracks and improved localisation for Whareformer.

\subsection{Cross-dataset generalisation}

We now evaluate all models considered in the previous section (\ie trained on EPIC-KITCHENS) on two other test sets to assess their generalisation.

\pgheader{Results on IT3DEgo.} To verify generalisation to non-kitchen activities, we evaluate Whareformer on IT3DEgo and provide the results in~\Cref{tab:cross_dataset_results} (middle).
Here, ByteTrack completely fails, unable to form consistent tracks with reliable 3D location estimates.
Furthermore, 3D-informed methods such as IT3DEgo and LMK struggle to generalise and exhibit large drops in performance: LMK experiences a 22.0\% absolute drop in mPCL compared to the dataset it was tuned on. In contrast, Whareformer remains strong across all metrics, achieving a significant +27.1\% improvement in mPCL and +18.7\% in IDF1 over LMK. The PCL plot in~\Cref{fig:cross_dataset_pcl} shows that Whareformer achieves better PCL over all timescales, verifying the model can generalise to diverse scenes.

\pgheader{Results on HD-EPIC.} We also evaluate on the challenging validation-only HD-EPIC. As shown in~\Cref{tab:cross_dataset_results}~(right), LMK and LMK-Inf exhibit a similar performance drop to the IT3DEgo dataset. ByteTrack achieves near-random results.
The IT3DEgo method benefits from its enrolment procedure: by registering objects at their first appearance, the method gains an advantage in HD-EPIC, particularly because object annotations are sparse. Having an oracle knowledge of the ground-truth number of tracks $K$ gives a significant advantage. In contrast, Whareformer does not use such privileged information, yet still outperforms this method. Examining the distinction between the ``where'' and the ``what'' reveals a clear advantage: Whareformer yields improvement in spatial localisation over LMK (+7.1\% mPCL), but notably delivers a more significant leap in track consistency (+23.6\% IDF1). 
The spatial localisation improvement is consistent across all timescales, as shown in \Cref{fig:cross_dataset_pcl}. 

\begin{figure}[t]
    \centering
    \includegraphics[width=\linewidth]{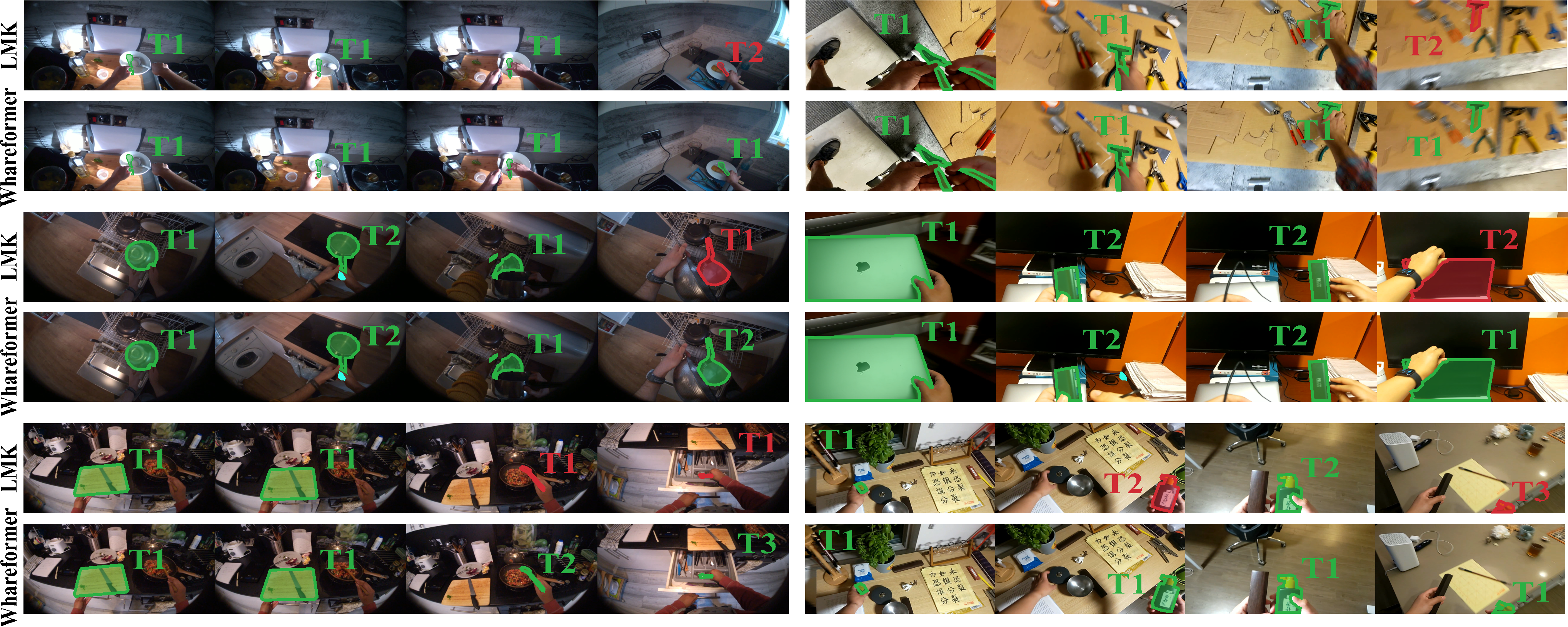}
    \caption{\textbf{Qualitative comparison for challenging scenarios.} We compare LMK and Whareformer on HD-EPIC (left) and IT3DEgo (right). \textbf{Top:} Whareformer maintains consistent track identities through pick-and-place motions (left) and motion blur (right). \textbf{Middle:} Whareformer successfully disambiguates spatially close objects without merging tracks. \textbf{Bottom:} Whareformer maintains precision in clutter (left) and sustains identity during long-distance room transits (right).}
    \label{fig:cross_dataset_qualitative_scenarios}
\end{figure}

\pgheader{Qualitative results.} \Cref{fig:cross_dataset_qualitative_scenarios} further highlights Whareformer's tracking robustness across HD-EPIC and IT3DEgo.
The first row demonstrates track continuity through a pick-and-place motion and ego-motion blur; Whareformer correctly connects a repositioned whisk and handles blurred observations of a clamp, whereas LMK fragments these into new tracks.
The second row illustrates precision with spatially close objects. Whareformer successfully distinguishes a saucepan placed exactly where a colander was previously located, and maintains separate identities for an interacting laptop and dongle, avoiding LMK's track merging.
The final row shows robust performance in clutter and long object transits. It accurately isolates three close items (chopping board, wooden spoon, and teaspoon) and perfectly follows an ink bottle carried across a room—scenarios where baseline methods suffer identity switches and trajectory fragmentation.

\subsection{Ablations}
\label{subsec:ablations}

In this section, we challenge a number of our design choices and assess their importance in Whareformer. Unless specified, ablations are run on the test set of the EPIC-KITCHENS dataset.

\pgheader{Core components.}
In~\Cref{tab:core_component_ablation} we analyse the impact of Whareformer's core components: the appearance information, location information, and the new track token.
Each component is individually ablated to assess its impact. 
Results show that all three components are required for achieving optimal performance.

Removing appearance information causes a collapse in both localisation and identity preservation (34.0\% mPCL, 50.5\% IDF1), confirming appearance as a critical signal as objects come close together as well as move and occupy the same location as a previous object.

Excluding the location also degrades performance (62.6\% mPCL, 79.5\% IDF1), showing its role in resolving ambiguities between visually similar objects or identical appearances of distinct instances (e.g.\ two forks).

To ablate the New Track token ($\mathrm{NT}$), we remove the token and create a new track only when the highest existing track assignment score is below 0.25. We also compare to a fixed, randomly initialised variant. Results show that disabling the $\mathrm{NT}$ token significantly reduces performance (53.0\% mPCL, 69.7\% IDF1), confirming that it is a crucial component of our model and a key contribution. Learning this token brings a small extra boost in performance.

\pgheader{Appearance representation.}
First, we demonstrate the effectiveness of our track representation strategies. The analysis focuses on the performance metrics, as well as the computational resources required.
To validate our choice of DenStream for the track appearance representation, we compare to alternatives.

As shown in~\Cref{tabe:representation_strategies} (top), DenStream achieves the highest performance across all metrics (72.2\% mPCL, and 84.0\% IDF1), while maintaining moderate memory usage (220.7 MB) and practical runtime (322 FPS). This balance of accuracy and efficiency makes it well-suited for online tracking. In contrast, short-term methods such as the most recent appearance, one-second window, and 100-frame moving average fail to capture the track's overall appearance leading to reduced tracking performance. Conversely, infinite-memory performs similarly but is prohibitively resource-intensive.

\input{tables/ablations}

\pgheader{Location representation.}
We validate our choice of a one-second temporal window and compare to the same strategies used for appearance, with results shown in~\Cref{tabe:representation_strategies} (bottom). All strategies perform comparably, however the one-second window achieves the best performance and remains scalable.
Overall, Whareformer is more sensitive to the choice of the appearance representation.

\pgheader{Assignment encoder design.} To evaluate the impact of our encoder design, we test models with no encoder, a linear encoder (a linear layer with ReLU), an MLP (two linear layers), and Transformer encoders (1 to 4 layers). \Cref{tab:layers_ablation} shows that all attention-based models outperform alternatives in both localisation and track consistency, demonstrating the benefit of attention across track assignments including the new token. Extra encoder layers lead to overfitting.

\pgheader{Depth alignment.} We also show the effect of depth alignment when obtaining 3D locations in~\Cref{tab:depth_alignment}. The results show that 
removing depth alignment leads to a modest performance drop, indicating robustness to \textit{using metric depth alone}.

\begin{wraptable}[10]{r}{0.53\textwidth}
    \caption{Impact of the training strategy.}
    \label{tab:dagger_ablation}
    \scriptsize
    \begin{tabular}{l rr rr}
        \toprule
        & \multicolumn{2}{c}{\textbf{Teacher Forcing}} 
        & \multicolumn{2}{c}{\textbf{DAgger Correction}} \\
        \cmidrule(lr){2-3} \cmidrule(lr){4-5}
        \textbf{Dataset} & mPCL & IDF1 & mPCL & IDF1 \\
        \midrule
        EPIC & \textbf{72.2} & \textbf{84.0} & 71.0 & 82.8 \\
        IT3DEgo & 58.1 & 87.9 & \textbf{58.6} & \textbf{88.5} \\
        HD-EPIC & 41.6 & 51.6 & \textbf{43.6} & \textbf{52.4} \\
        \bottomrule
    \end{tabular}
\end{wraptable}
\pgheader{DAgger correction.} Whareformer is trained using teacher forcing, where tracks are constructed from ground-truth assignments. However, during inference, tracks are formed based on Whareformer's own predictions, causing a distribution shift. To address this, we ablate training Whareformer using DAgger-style correction~\cite{ross2011dagger}, re-extracting training data every 30 epochs using Whareformer's predicted assignments. Batches are formed by sampling from ground-truth and predicted data, with increasing probability for the latter as training progresses. Results in~\Cref{tab:dagger_ablation} show that DAgger improves generalisation on IT3DEgo and HD-EPIC.

\pgheader{Additional ablations.}
We report further ablations in the supplementary material to verify design choices related to PCA and DenStream configurations.

%% file: tables/main_results.tex
\begin{figure*}[t!]
    \centering
    \begingroup
    \captionof{table}{\textbf{Comparisons across datasets}.
    The same model, trained on EPIC-KITCHENS, is directly applied to all three test sets: the corresponding EPIC-KITCHENS test set, as well as IT3DEgo and HD-EPIC.
    \textcolor{gray}{$^*$ IT3DEgo} enrols the first observation of each object before tracking and assumes a known number of objects $M=K$, gaining an advantage. $^{\dagger}$ OSOM oracle perfectly tracks objects while they are in-view so IDF1 is trivially \textit{100} but ignores them for when out-of-view.
    }
    \label{tab:cross_dataset_results}
    \scriptsize
    \begin{tabular*}{\textwidth}{@{\extracolsep{\fill}} l rrrrrr @{}}
    \toprule
    {} & \multicolumn{2}{c}{\textbf{EPIC-KITCHENS}} 
       & \multicolumn{2}{c}{\textbf{IT3DEgo}} 
       & \multicolumn{2}{c}{\textbf{HD-EPIC}} 
       \\ \cmidrule{2-7}
        
        \textbf{Method} 
        & \textbf{mPCL} & \textbf{IDF1}
        & \textbf{mPCL} & \textbf{IDF1}
        & \textbf{mPCL} & \textbf{IDF1} \\
        \midrule
        
        OSL~\cite{Plizzari2025OSNOM}
        & 1.2 & 36.8
        & 1.1 & 21.4
        & 0.7 & 41.9 \\
    
        Random Matching
        & 2.6 & 1.4
        & 2.5 & 3.3
        & 21.2 & 15.2 \\

        OSOM$^{\dagger}$~\cite{Plizzari2025OSNOM}
        & 8.3 & \textit{(100)}
        & 26.5 & \textit{(100)}
        & 0.7 & \textit{(100)} \\
    
        ByteTrack~\cite{zhang2022bytetrack}
        & 12.3 & 31.0
        & 0.8  & 7.5
        & 21.8 & 28.8 \\
    
        \textcolor{gray}{ IT3DEgo*~\cite{zhao2024instance}}
        & \textcolor{gray}{40.0} & \textcolor{gray}{71.0}
        & \textcolor{gray}{27.5} & \textcolor{gray}{57.1}
        & \textcolor{gray}{39.2} & \textcolor{gray}{51.0}  \\
        
        LMK~\cite{Plizzari2025OSNOM}
        & 53.0 & 70.0
        & 31.0 & 69.2
        & 34.5 & 28.0 \\
        
        LMK-Inf 
        & 60.9 & 76.6
        & 40.8 & 75.5
        & 36.5 & 31.9 \\ \midrule
    
        Whareformer
        & \textbf{72.2} & \textbf{84.0}
        & \textbf{58.1} & \textbf{87.9}
        & \textbf{41.6} & \textbf{51.6} \\
        
        ~~\textit{Abs.~gain vs. LMK~\cite{Plizzari2025OSNOM}}
        & \scriptsize\textcolor{ForestGreen}{\emph{\textbf{$\uparrow$19.2\%}}} 
        & \scriptsize\textcolor{ForestGreen}{\emph{\textbf{$\uparrow$14.0\%}}} 
        & \scriptsize\textcolor{ForestGreen}{\emph{\textbf{$\uparrow$27.1\%}}} 
        & \scriptsize\textcolor{ForestGreen}{\emph{\textbf{$\uparrow$18.7\%}}} 
        & \scriptsize\textcolor{ForestGreen}{\emph{\textbf{$\uparrow$7.1\%}}} 
        & \scriptsize\textcolor{ForestGreen}{\emph{\textbf{$\uparrow$23.6\%}}} \\
        \bottomrule
    \end{tabular*}
    \endgroup
    \nextfloat%
    \input{plots/pcl_plots}%
    \caption{\textbf{Comparisons across datasets}. PCL over sequence length (time).
    We compare to LMK~\cite{Plizzari2025OSNOM}, the SOTA on EPIC. Whareformer consistently outperforms LMK.
    }
    \label{fig:cross_dataset_pcl}
\end{figure*}

%% file: plots/pcl_plots.tex
\begin{tikzpicture}
    \begin{groupplot}[
        group style={
            group size=3 by 1,
            horizontal sep=0.8cm,
        },
        scale only axis=true,
        width=0.269\linewidth,
        height=2.2cm,
        xmin=0, xmax=13,
        ymin=5, ymax=100,
        xtick={1,2,4,8,12},
        grid=both,
        minor tick num=3,
        every axis plot/.append style={line width=.8pt, mark size=1.5pt},
        title style = {font=\footnotesize, yshift=-4pt, text height=2.5ex, text depth=0.25ex, inner sep=1.5pt},
        xlabel style = {font=\scriptsize, yshift=-2pt, inner sep=0pt},
        ylabel style = {font=\scriptsize, yshift=-3pt, inner sep=0pt},
        xticklabel style = {font=\scriptsize, inner sep=2.0pt},
        yticklabel style = {font=\scriptsize, inner sep=2.0pt},
        xlabel={Time (min)},
        ylabel={PCL (\%)}
    ]

    \nextgroupplot[title={EPIC-KITCHENS}]
        \addplot[color=BrickRed, mark=triangle*] coordinates {
            (0.00,100.0)
            (0.33,86.64)
            (0.66,84.01)
            (1.00,81.75)
            (2.00,77.21)
            (4.00,72.79)
            (8.00,68.84)
            (12.0,65.36)
        };
        \addplot[color=black, mark=*] 
        coordinates {
            (0.00,100.0)
            (0.33,76.52)
            (0.66,71.77)
            (1.00,68.54)
            (2.00,62.98)
            (4.00,56.18)
            (8.00,49.96)
            (12.0,39.88)
        };

    \nextgroupplot[title={IT3DEgo}]
        \addplot[color=BrickRed, mark=triangle*] coordinates {
            (0.00,100.0)
            (0.33,84.73)
            (0.66,80.77)
            (1.00,77.69)
            (2.00,69.77)
            (4.00,59.70)
            (8.00,47.18)
            (12.0,46.70)
        };
        \addplot[color=black, mark=*] 
        coordinates {
            (0.00,100.0)
            (0.33,77.38)
            (0.66,70.13)
            (1.00,64.51)
            (2.00,50.04)
            (4.00,38.81)
            (8.00,18.97)
            (12.0,10.87)
        };

    \nextgroupplot[
        title={HD-EPIC},
        legend pos=north east,
        legend columns=1,
        legend style={
            font=\scriptsize,
            draw=black,
            fill=white,
            at={(0.98,0.98)},
            anchor=north east,
            inner sep=1.5pt %
        }
    ]
        \addplot[color=BrickRed, mark=triangle*] coordinates {
            (0.00,100.0)
            (0.33,59.73)
            (0.66,54.10)
            (1.00,51.39)
            (2.00,47.27)
            (4.00,41.98)
            (8.00,36.79)
            (12.0,34.73)
        };
        \addlegendentry{Whareformer}

        \addplot[color=black, mark=*]
        coordinates {
            (0.00,100.0)
            (0.33,52.15)
            (0.66,45.60)
            (1.00,42.16)
            (2.00,37.11)
            (4.00,34.61)
            (8.00,30.82)
            (12.0,28.98)
        };
        \addlegendentry{LMK~\cite{Plizzari2025OSNOM}}
    \end{groupplot}
\end{tikzpicture}

%% file: tables/ablations.tex
\begin{table}[t]
    \centering
    \begin{minipage}[t]{0.39\linewidth}
        \centering
        \captionof{table}{Effect of some of Whareformer's core components: appearance ($A$), location ($L$), and the $\mathrm{NT}$ token.}
        \label{tab:core_component_ablation}
        \scriptsize
        \begin{tabular}[t]{l c c c c c c}
            \toprule
            \textbf{Ablation} & $A$ & $L$ & $\mathrm{NT}$ & \textbf{mPCL} & \textbf{IDF1} \\
            \midrule
            $L$-Only & \xmark & \cmark & \cmark & 34.0 & 50.5 \\
            No $\mathrm{NT}$ & \cmark & \cmark & \xmark & 53.0 & 69.7 \\
            $A$-Only & \cmark & \xmark & \cmark & 62.6 & 79.5 \\
            \midrule
            Fixed $\mathrm{NT}$ & \cmark & \cmark & (\cmark) & 72.0 & 83.9 \\
            \rowcolor{LightGrey} \textit{Whareformer} & \cmark & \cmark & \cmark & \textbf{72.2} &\textbf{84.0} \\
            \bottomrule
        \end{tabular}
    \end{minipage}%
    \hfill
    \begin{minipage}[t]{0.59\linewidth}
        \centering
        \captionof{table}{Comparison of track appearance and location representation strategies.}
        \label{tabe:representation_strategies}
        \scriptsize
        \begin{tabular}[t]{l c c r c}
            \toprule
            \textbf{Representation} & \textbf{mPCL} & \textbf{IDF1} & \textbf{MB} & \textbf{Avg. FPS} \\
            \midrule  
            \multicolumn{5}{l}{\textit{Appearance} (with 1 sec window location)} \\
            \midrule  
            Most recent appearance & 52.9 & 69.7 & \textbf{1.4} & \textbf{798} \\
            1 sec window & 59.4 & 74.7 & 76.7 & 603 \\
            100 moving avg. & 55.5 & 70.7 & 137.4 & 580 \\
            \rowcolor{LightGrey} \textit{DenStream} & \textbf{72.2} & \textbf{84.0} & 220.7 & 322  \\
            Infinite memory & 70.6 & 83.2 & 1843.8 & 50 \\ 
            \midrule
            \multicolumn{5}{l}{\textit{Location} (with DenStream appearance)} \\
            \midrule  
            Most recent location & 69.7 & 83.3 & \textbf{$\approx$0.0} & 307 \\
            \rowcolor{LightGrey} \textit{1 sec window} & \textbf{72.2} & \textbf{84.0} & 0.9 & \textbf{322} \\ 
            100 moving avg. & 68.3 & 82.3 & 1.5 & 299 \\
            DenStream & 70.3 & 83.8 & 0.1 & 311 \\
            Infinite memory & 70.3 & 83.7 & 21.6 & 260 \\
            \bottomrule
        \end{tabular}
    \end{minipage}
    
    \begin{minipage}[b]{0.39\linewidth}
        \centering
        \captionof{table}{Effect of the assignment encoder design.}
        \label{tab:layers_ablation}
        \scriptsize
        \begin{tabular}[b]{l c c}
            \toprule
            \textbf{Encoder Design} & \textbf{mPCL} & \textbf{IDF1} \\
            \midrule
            No encoder & 66.6 & 81.2 \\
            Linear encoder & 67.3 & 81.3 \\
            MLP encoder & 67.6 & 82.0 \\
            \rowcolor{LightGrey} \textit{1 encoder layer} & \textbf{72.2} & \textbf{84.0} \\
            2 encoder layers & 70.8 & 83.2 \\
            3 encoder layers & 70.7 & 83.1 \\
            4 encoder layers & 71.0 & 83.5 \\
            \bottomrule
        \end{tabular}
    \end{minipage}%
    \hfill
    \begin{minipage}[b]{0.59\linewidth}
        \centering
        \captionof{table}{Effect of using depth alignment to obtain location estimates.}
        \label{tab:depth_alignment}
        \scriptsize
        \begin{tabular}[b]{l rr rr}
            \toprule
            & \multicolumn{2}{c}{\textbf{No Alignment}}  
            & \multicolumn{2}{c}{\textbf{With Alignment}} \\
            \cmidrule(lr){2-3} \cmidrule(lr){4-5}
            \textbf{Method} & mPCL & IDF1 & mPCL & IDF1 \\
            \midrule
            LMK & 52.4 & 69.4 & 53.0 & 70.0 \\
            LMK-Inf & 60.2 & 75.9 & 60.9 & 76.6 \\
            Whareformer & 70.5 & 83.1 & \textbf{72.2} & \textbf{84.0} \\
            \bottomrule
        \end{tabular}
    \end{minipage}
\end{table}

%% file: sec/5_conclusion.tex
\section{Conclusion}
\label{sec:conclusion}

We introduced \textit{Whareformer}, a transformer-based model for long-term 3D object tracking in egocentric videos. The model learns to associate current observations with existing tracks or create new ones by jointly reasoning over appearance and location. This approach maintains consistent representations even when objects are out of view. 
\textit{Whareformer} achieves significant improvements on the OSNOM task, generalising across diverse egocentric datasets, with novel objects and scenes. It benefits from a few innovations including a learnt token for new tracks, jointly reasoning over distances between observations and tracks rather than actual features, and a robust appearance representation of the tracks to accommodate objects changing appearance during interactions.
These innovations allow our transformer-based model, \textit{Whareformer}, to achieve strong performance using a relatively small training set.

\pgheader{Acknowledgements.} This work uses public datasets. Research at Bristol is supported by EPSRC UMPIRE EP/T004991/1, EPSRC PG Visual AI EP/T028572/1. J Chalk and S Sinha are supported by ESPRC Doctoral Training Program (DTP). 
We acknowledge the usage of GPU Node hours provided by the Isambard-AI National AI Research Resource (AIRR) granted as part of the AIRR Gateway project ``HOI Foundational Model from Egocentric Data'' (Dec 2025 - Mar 2026) and the Sovereign AI Unit call project
``Gen Model in Ego-sensed World'' (Aug 2025 - Nov 2025).

%% file: sec/X_suppl.tex
\newpage
\appendix

\section{Additional qualitative examples}
We provide video-based qualitative results for all evaluated datasets on our project webpage: \url{https://jacobchalk.github.io/Whareformer/}. 

For EPIC-KITCHENS-100, we provide a qualitative sample illustrating complex, multi-object interactions and tracking. We also supply videos showcasing IT3DEgo's diverse, unseen environments and for HD-EPIC's sparse, difficult sequences. Each video is shown in a side-by-side format, with LMK \cite{Plizzari2025OSNOM} on the left (state-of-the-art for the OSNOM task) and Whareformer on the right, enabling direct comparison of their tracking and 3D localisation behaviour. For visualisation, we colour an object's bounding box \textcolor{Green}{green} if the current observation has the same track ID as it had for the object's first observation. Otherwise, we  colour the bounding box \textcolor{Orange}{orange} indicating that an ID switch has occurred, but the original track can still localise the object within a 30cm threshold (corresponding to the distance threshold for the PCL evaluation). However, if the original track cannot localise the object within this threshold, we colour the box \textcolor{Red}{red}, capturing both ``what'' and ``where'' errors. Across all datasets, Whareformer demonstrates fewer errors and  greater temporal stability than LMK.

\section{DenStream algorithm details}
DenStream~\cite{cao2006density} is an online density-based clustering method designed to handle evolving data streams; here, we describe how we adapt it to our object tracking setting, rather than presenting the original algorithm. Specifically, we detail three aspects of the algorithm: how appearance information is modelled, the parameters affecting the representation, and how micro-clusters are pruned.

\pgheader{Appearance modelling with DenStream.} DenStream maintains a dynamic summary of the appearance data stream through persistent micro-clusters (\textit{p-micro-clusters}) and transient micro-clusters (\textit{t-micro-clusters}). We note that the distinction between these two types of micro-clusters only affects the pruning strategy (detailed following the parameters); it does not change the way appearance distances between observations and tracks ($c^{A}_{n,k}$) are computed. 

Conceptually, p-micro-clusters represent stable, recurring appearances, while t-micro-clusters capture transient or newly emerging views. Each new appearance feature is processed in a single pass: it is first checked against existing p-micro-clusters and merged into the closest one if a suitable match exists; otherwise, it is compared to t-micro-clusters or used to create a new t-micro-cluster if still deemed unsuitable. A temporal decay prioritises recent appearances while retaining older, less frequent ones with reduced influence, enabling adaptation to changing appearance and recovery when revisiting past viewpoints. Notably, DenStream does not require an offline calibration stage like alternative algorithms such as StreamKM++~\cite{ackermann2012streamkm++}, making it well-suited to online tracking.

\pgheader{Parameters.} The DenStream appearance representations are influenced by three parameters: $\epsilon$ determines the maximum radius of a micro-cluster---if adding an observation to its nearest micro-cluster forces the clusters radius to exceed this threshold, a new t-micro-cluster is created instead. $\mu$ is a threshold to determine when a t-micro-cluster is upgraded to a p-micro-cluster, affecting the way it is pruned in future frames.\footnote{The original DenStream paper has two separate parameters $\beta$ and $\mu$ which are multiplied. For brevity, we combine them to a single parameter.} Finally, $\lambda$ is the temporal decay and controls the influence of old clusters, with larger values leading to faster forgetting. A temporal pruning threshold $\tau_{p}$ is defined by $\mu$ and $\lambda$ as: 

\begin{equation}
    \tau_{p}=\left\lceil \frac{1}{\lambda}\ln\frac{\mu}{\mu-1}\right\rceil
\end{equation}

This threshold triggers an update of the micro-clusters once a track has not been observed for a sufficient period, which we detail next.

\pgheader{Pruning micro-clusters.} All micro-clusters are associated with a weight, representing the cumulative presence of the cluster over time (\ie how often and how recently it has been observed). When tracks are updated, the weight of each micro-cluster decays exponentially at time $t$ as

\begin{equation}
    w(t) = w(t_{u}) \cdot 2^{-\lambda (t - t_{u})},
\end{equation}

where $t_{u}$ is the micro-clusters previous update time and $w(0) = 1$. During the pruning stage, a p-micro-cluster is retained only if $w(t)\geq \mu$, while a t-micro-cluster created at time $t_{c}$ is retained if: 

\begin{equation}
    w(t)\;\geq\;\frac{2^{-\lambda (t - t_{c} + \tau_{p})}-1}{2^{-\lambda \tau_{p}}-1}.
\end{equation}

When a track is updated, we check the elapsed time since its last update; if this duration exceeds the pruning threshold $\tau_{p}$, the appearance representation is pruned, triggering a refreshed update. If sufficient time has not passed, pruning is skipped to avoid compromising the currently updating representation and affecting assignment decisions in the immediate, subsequent frames. This strategy ensures that only resumed tracks---those that have experienced a significant temporal gap---undergo pruning, preserving the integrity of tracked objects in the current frame(s) while still allowing the appearance memory to adapt over long-term tracking.

In addition to pruning, a t-micro-cluster is promoted to a p-micro-cluster whenever its weight rises above $\mu$, which it must remain above to avoid removal in future pruning cycles. If no micro-clusters remain after pruning, the cluster with the largest weight is preserved and demotes to a t-micro-cluster in order to maintain stability and ensure the track's appearance is not completely forgotten.

\section{Ground-truth creation}

To construct the actual training data from the set of EPIC-KITCHENS-100 training videos, 
we first obtain perfect tracking information by running an oracle tracker, which assigns observations to tracks based on ground-truth identities. For each observation in a frame, we store the track features $a'$ and $l'$ that yield the minimum appearance distance $c^{A}_{n,k}$ and location distance $c^{L}_{n,k}$. This is performed for all currently existing tracks at time $t$, producing a feature tensor $\mathbf{I}^{t} \in \mathbb{R}^{O^{t} \times (T^{t}+1) \times (D+3)}$ where $O^{t}$ is the number of observations at time $t$ and $T^{t}$ is the number of existing tracks.

Each of the $T^{t}+1$ entries along the second dimension corresponds either to an existing track or to the observation itself. The final $D+3$ dimension contains the $D$-dimensional appearance descriptor together with the 3D location. Therefore, this tensor provides, for every observation, a complete set of pairwise appearance--location features relative to all existing tracks at time $t$.

The model takes $\mathbf{I}^{t}$ as input and produces a distance matrix, containing the minimum appearance distances $c^{A}_{n,k}$ and location distances $c^{L}_{n,k}$ for every observation--track pair before producing the embedded track assignment likelihoods.

In addition to $\mathbf{I}^{t}$, we store a one-hot vector $\mathbf{y}^{t} \in \{0,1\}^{T^{t}+1}$ for each observation indicating the ground-truth assignment. This supervision signal is used during training to specify whether the observation should be matched to one of the existing tracks or initiate a new track.

\section{Implementation details}
\label{subsec:implementation_details-whareformer}

This section details the implementation of Whareformer, outlining its architectural, training, and appearance representation details. 

\noindent\textbf{Architectural details.} The core of our model is a single layer Transformer encoder using $d=64$. The layer utilises 32 attention heads and a feed-forward network with a ReLU activation and an inner dimension of $4d=256$. We employ a pre-layer normalisation configuration within the encoder block, as we found this to be more stable, and apply layer normalisation and dropout to the output of the encoder. The input embedding network $g(\cdot)$, which converts each of the appearance-location distance pairs into the embedded track assignment likelihoods, consists of a 2-to-$d$ linear layer followed by a ReLU activation, layer normalisation, and a dropout layer. The classification head $h(\cdot)$ is a simple $d$-to-1 linear layer. A dropout rate of $p=0.1$ is applied to, the output of the embedding network, within the encoder layer, and to the output of the encoder.

\noindent\textbf{Training details.} We train the model for 100 epochs using the AdamW optimiser~\cite{loshchilov2017decoupled} with a batch size of 256 and a learning rate of $1\mathrm{e}{-4}$ and weight decay of $5\mathrm{e}{-3}$. We employ a linear warm-up for the first 5 epochs, increasing the learning rate from $1\mathrm{e}{-6}$ to its target value, followed by a cosine decay schedule for the remainder of training.

Because the number of existing tracks varies across frames, the second dimension is padded to the maximum number of tracks present within the batch, and masking is applied so that padded feature entries do not contribute to the model's predictions or loss. This enables stable batching while preserving the integrity of the original ground-truth supervision.

\noindent\textbf{Appearance representations.} To capture the appearance of each observation, we extract visual features using the DINOv2 model~\cite{oquab2024dinov} with a ViT-g backbone configured without register tokens, using the CLS token as the representation, following~\cite{Plizzari2025OSNOM}. The resulting 1536-D visual features are then  $\ell_{2}$-normalised, projected to a $D=256$ space via a learned PCA transformation, and normalised again before being used to compute the appearance distance $c^{A}_{n,k}$. For the DenStream parameters, we set $\epsilon=0.25$, $\mu=10.0$ and  $\lambda=1\mathrm{e}{-3}$, giving a pruning threshold $\tau_{p}\approx106$ seconds.

\section{Further ablations}

We perform additional ablations on the architectural design and appearance representations for observations and tracks. For all ablations, we select the model with the highest mPCL and IDF1.

\input{tables/supplementary_results}

\pgheader{Size of hidden dimension.} We investigate the hidden dimension of the likelihood embedding vectors (\Cref{tab:hidden_ablation}). All configurations yield similarly consistent tracks (IDF1), but localisation accuracy (mPCL) varies with hidden dimension. The chosen dimension of 64 shows clear improvements for localisation and identity preservation.

\pgheader{Number of transformer heads.} We analyse the impact of the number of attention heads in our Transformer encoder, varying the count from 1 to 32. The results in~\Cref{tab:attn_heads_ablation} show that the model is robust across all configurations, with minimal changes in performance.

\pgheader{Appearance feature type.} We compare how the method of obtaining appearance features affects results. We compare providing the model with a standard bounding box crop (\ie, including background pixels) vs.\ applying a mask (ground truth or predicted by SAM2~\cite{ravi2025sam}) to the cropped region. We see that applying a mask yields more discriminative features and a significant performance boost over the unmasked bounding box. Whareformer clearly scales with the input quality, improving significantly with predicted SAM masks and peaking when using ground-truth masks.

\pgheader{Visual feature dimensionality.} We ablate how to generate the most discriminative appearance distances for the model to learn from. We investigate how reducing the raw 1536-D DINOv2 features using PCA affects the final model performance and compare against the raw features themselves, which are $\ell_{2}$-Normalised for stability. The results are shown in~\Cref{tab:pca_ablation}. Overall, only the 32-D PCA configuration shows a noticeable drop, indicating that over-reduction of the features limits discriminative power. We note that the model can operate effectively using the full 1536-D raw features.

\pgheader{DenStream configuration.} We ablate the key hyperparameters of DenStream that govern how a track's appearance is summarised over time: the maximum micro-cluster radius ($\epsilon$), the persistent micro-cluster threshold ($\mu$), and the temporal decay rate ($\lambda$). Results are shown in~\Cref{tab:denstream_ablation}. The radius $\epsilon$ has the strongest effect: too small ($\epsilon=0.1$) yields overly specific clusters that are quickly pruned, while too large ($\epsilon=0.5$) merges distinct appearances, reducing discriminability. The parameters $\mu$ and $\lambda$ control the trade-off between memory and adaptability, and the model is generally robust to their values.

\pgheader{Conflict resolution.}
We evaluate three strategies for resolving conflicting assignments using the model's softmax scores. Our chosen \emph{greedy} strategy assigns observations in descending confidence order, always accepting new-track decisions and locking tracks once assigned. A \emph{Hungarian} strategy performs optimal matching on an augmented matrix that enables multiple new-track decisions. An \emph{iterative} strategy has the model repeatedly reprocess the least confident conflicting assignments until all observations are assigned. The results in~\Cref{tab:conflict_strategy} show that the conflict resolution strategy has minimal impact on performance, hence we select the cheapest computationally (greedy).

%% file: tables/supplementary_results.tex
\begin{table}[t]
    \centering
    \begin{minipage}[t]{0.49\linewidth}
        \centering
        \captionof{table}{Effect of changing the size of the hidden dimension in Whareformer. The model uses 1 encoder layer and 32 attention heads. The highlighted row is our selected model.}
        \label{tab:hidden_ablation}
        \scriptsize
        \begin{tabular}{c c c}
            \toprule
            \textbf{Hidden Dim} & \textbf{mPCL} & \textbf{IDF1} \\
            \midrule 
            32 & 71.2 & 82.8 \\
            \rowcolor{LightGrey} 64 &  \textbf{72.2}& \textbf{84.0} \\
            128 & 69.9 & 83.0 \\
            256 & 68.9 & 82.7 \\
            512 & 68.1 & 81.8 \\
            1024 & 70.5 & 82.3 \\
            \bottomrule
        \end{tabular}
    \end{minipage}%
    \hfill%
    \begin{minipage}[t]{0.49\linewidth}
        \centering
        \captionof{table}{Effect of changing the size of the PCA dimension when calculating the appearance distance as input to Whareformer. *: Indicates the raw DINOv2 visual features that are $\ell_{2}$-normalised.}
        \label{tab:pca_ablation}
        \scriptsize
        \begin{tabular}{c c c c}
            \toprule
            \textbf{PCA Dim} & \textbf{mPCL} & \textbf{IDF1} \\
            \midrule 
            32 & 65.4 & 80.7 \\
            64 & 68.2 & 82.0 \\
            128 & 70.1 & 83.6 \\
            \rowcolor{LightGrey} 256 & \textbf{72.2} & \textbf{84.0} \\
            512 & 70.8 & 83.3 \\
            1536* & 70.8 & 82.5 \\
            \bottomrule
        \end{tabular}
    \end{minipage}
    \bigskip%
    \begin{minipage}[t]{0.49\linewidth}
        \centering
        \captionof{table}{Effect of changing the number of attention heads in Whareformer. The number of encoder layers is fixed to 1 and the highlighted row is our selected model.}
        \label{tab:attn_heads_ablation}
        \scriptsize
        \begin{tabular}{c c c}
            \toprule
            \textbf{\# Head} & \textbf{mPCL} & \textbf{IDF1} \\
            \midrule 
            1 & 71.8 & 83.0 \\
            2 & 71.0 & 82.7 \\
            4 & 70.7 & 82.5 \\
            8 & 71.0 & 83.1 \\
            16 & 71.2 & 83.3 \\
            \rowcolor{LightGrey} 32 & \textbf{72.2} & \textbf{84.0} \\
            \bottomrule
        \end{tabular}
    \end{minipage}%
    \hfill%
    \begin{minipage}[t]{0.49\linewidth}
        \centering
        \captionof{table}{Effect of the DenStream configuration: $\epsilon$, $\mu$, and $\lambda$ affecting appearance representations of tracks. The highlighted row is our selected configuration.}
        \label{tab:denstream_ablation}
        \scriptsize
        \begin{tabular}{ c c c c c}
            \toprule
            $\epsilon$ & $\mu$ & $\lambda$ & \textbf{mPCL} & \textbf{IDF1} \\
            \midrule
            0.1 & 10.0 & $1\mathrm{e}{-3}$ & 67.6 & 81.3 \\
            0.5 & 10.0 & $1\mathrm{e}{-3}$ & 67.0 & 80.4 \\
            0.25 & 5.0 & $1\mathrm{e}{-3}$ & 68.7 & 82.2 \\
            0.25 & 20.0 & $1\mathrm{e}{-3}$ & 70.5 & 83.4 \\
            0.25 & 10.0 & $1\mathrm{e}{-4}$ & 71.0 & 83.3 \\
            0.25 & 10.0 & $1\mathrm{e}{-2}$ & 69.1 & 81.5 \\
            \midrule
            \rowcolor{LightGrey} 0.25 & 10.0 & $1\mathrm{e}{-3}$ & \textbf{72.2} & \textbf{84.0} \\ 
            \bottomrule
        \end{tabular}
    \end{minipage}
    \bigskip%
    \begin{minipage}[b]{0.49\linewidth}
        \centering
        \captionof{table}{Effect of using bounding box, predicted, or ground truth mask features.}
        \label{tab:object_crop_ablation}
        \scriptsize
        \begin{tabular}{l c c}
            \toprule
            \textbf{Feature Type} & \textbf{mPCL} & \textbf{IDF1} \\
            \midrule 
            Bounding box & 49.4 & 66.5 \\
            Predicted Mask & 62.9 & 78.6 \\
            \rowcolor{LightGrey} Ground Truth Mask & \textbf{72.2} & \textbf{84.0} \\
            \bottomrule
        \end{tabular}
    \end{minipage}%
    \hfill
    \begin{minipage}[b]{0.49\linewidth}
        \centering
        \captionof{table}{Effect of different conflict resolution strategies during inference.}
        \label{tab:conflict_strategy}
        \scriptsize
        \begin{tabular}{l c c}
            \toprule
            \textbf{Strategy} & \textbf{mPCL} & \textbf{IDF1} \\
            \midrule 
            Iterative & \textbf{72.2} & 83.6 \\
            Hungarian & \textbf{72.2} & \textbf{84.0} \\
            \rowcolor{LightGrey} Greedy & \textbf{72.2} & \textbf{84.0} \\
            \bottomrule
        \end{tabular}
    \end{minipage}

\end{table}